\documentclass[10pt,twocolumn,letterpaper]{article}
\usepackage[textwidth=6.875in,textheight=8.875in,columnsep=.3125in]{geometry}
\usepackage{times,amsmath,amssymb}
\usepackage[table,dvipsnames]{xcolor}
\usepackage{booktabs,graphicx,multirow,tabularx,makecell,adjustbox}
\usepackage{algorithm,algorithmic,enumitem}
\usepackage[font=small,labelsep=period]{caption}
\usepackage{subcaption}
\usepackage[numbers,sort&compress]{natbib}
\definecolor{discobg}{RGB}{232,240,252}
\definecolor{linkblue}{rgb}{0.21,0.49,0.74}
\usepackage[colorlinks,allcolors=linkblue]{hyperref}
\hypersetup{pdftitle={DiSCO: Defending Text-to-Image Generation through Distribution-Guided Contrastive Suffix Optimization},pdfauthor={Tong Zhang, Motasem Alfarra, Carlos Hinojosa, Christos Louizos, Bernard Ghanem}}
\newcommand{\ms}[2]{$#1_{\pm #2}$}
\newcommand{\bms}[2]{$\mathbf{#1}_{\pm #2}$}
\newcommand{\equalcontrib}{\textsuperscript{$\dagger$}}
\title{DiSCO: Defending Text-to-Image Generation through Distribution-Guided Contrastive Suffix Optimization}
\author{
 Tong Zhang\textsuperscript{1} \quad
 Motasem Alfarra\textsuperscript{2}\equalcontrib \quad
 Carlos Hinojosa\textsuperscript{1}\equalcontrib \\
 \textbf{Christos Louizos}\textsuperscript{2} \quad
 \textbf{Bernard Ghanem}\textsuperscript{1} \\[0.6em]
 \small\textsuperscript{1} King Abdullah University of Science and Technology (KAUST)\\
 \small\textsuperscript{2}\textcolor{black}{Qualcomm AI Research} \\[0.4em]
 \small\texttt{\{tong.zhang.1, carlos.hinojosa, bernard.ghanem\}@kaust.edu.sa} \\
 \small\texttt{\{malfarra, clouizos\}@qti.qualcomm.com} \\
 \small $\dagger$ Equal Contribution
}
\date{}
\begin{document}
\maketitle
\begin{abstract}
As text-to-image generative models advance, they raise critical safety concerns, particularly the generation of Not-Safe-For-Work (NSFW) content such as violence and nudity, further exacerbated by red-teaming adversarial attacks. Existing defenses predominantly operate under white-box assumptions, relying on text encoder optimization, weight editing, or inference-time intervention, and cannot scale to proprietary models. Moreover, existing defense systems still leave \textit{residual adversarial outputs}: harmful generations that bypass their safeguards, rooted in the model's learned output distribution. We propose DiSCO, a training-free, strictly black-box, plug-and-play defense that mitigates such residual adversarial outputs by contrastively optimizing their corresponding prompts, steering the generation distribution toward the safe region and away from the unsafe one, complementing the base model and existing defenses. DiSCO performs distribution-guided suffix expansion through fixed-length beam search, using contrastive scores against a few-shot reference set of only 25 safe and 25 unsafe images generated by the target system. Across four text-to-image architectures, four defenses, and four attacks, DiSCO reduces average NudeNet Attack Success Rate (ASR) from $32.9\%$ to $18.8\%$, with relative reductions of $37.3\%$ on undefended models and $52.5\%$ when combined with existing defenses, and generalizes to unseen prompt distributions. These results establish DiSCO as a black-box, architecture-agnostic safety module that complements existing defenses across text-to-image systems.
 \\\textcolor{red}{\textbf{Warning}: \textit{This paper contains model outputs that are offensive in nature.}}
\end{abstract}

\section{Introduction}
The onset of text-to-image (T2I) generation has transformed creative workflows across industries~\citep{liao2025imagegen, ramesh2021zero, Luo_2025_ICCV, nichol2021improved, dhariwal2021diffusion, song2019generative, ho2020denoising}, enabling users to synthesize photorealistic imagery from natural language descriptions. Models such as Stable Diffusion~\citep{rombach2022high}, SD 3~\citep{esser2024scaling}, and Flux~\citep{flux2024} have demonstrated remarkable generative capabilities, yet this progress has simultaneously introduced critical safety vulnerabilities. In the realm of content safety, these models can produce Not-Safe-For-Work (NSFW) content~\citep{li2024safegen, chen2025comprehensive}, including depictions of violence, nudity, and other harmful imagery, either through undefended generation or through adversarial attacks that deliberately manipulate the model's inputs or parameters. The growing accessibility of open-source T2I models has further amplified these risks, as users can freely interact with systems that lack adequate safety guardrails, raising urgent concerns for responsible deployment.

The discovery of red-teaming adversarial attacks has accordingly led to escalating security concerns surrounding T2I systems. These attacks span a spectrum of threat models, from black-box methods that craft adversarial prompts without model access~\citep{tsai2024ring, yang2024sneakyprompt}, to white-box approaches that leverage gradients and internal representations for targeted prompt optimization~\citep{zhang2024generate, yang2024mma}. More recently, Large Language Model (LLM)-driven attacks, such as APT~\citep{liu2025autoprompt}, have raised the bar further by generating human-readable adversarial prompts that evade both automated detectors and human review. The increasing sophistication of these attacks demands defense mechanisms that are equally robust and adaptive.

\begin{figure}[t]
  \centering
  \includegraphics[width=0.98\linewidth]{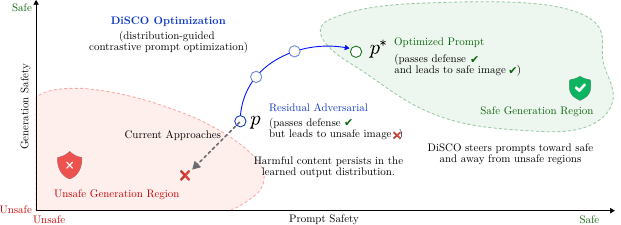}
\caption{Existing defenses leave \textit{residual adversarial outputs}: prompts $p$ that pass the defense yet still lead to unsafe generations, as harmful content persists in the model's learned output distribution. DiSCO optimizes $p$ into $p^*$ via a distribution-guided contrastive objective, steering generations toward the safe region and away from the unsafe one without modifying the system.}
  \label{fig:intro}
  \vspace{-8pt}
\end{figure}

In response, a spectrum of defenses has emerged. White-box methods intervene inside the generation pipeline via weight fine-tuning~\citep{gandikota2023erasing, zhang2024defensive}, cross-attention editing~\citep{gong2024rece}, or inference-time steering~\citep{schramowski2023safe, yoon2025safree}, but they inherently do not scale to closed-source models. Prompt-level defenses offer a lighter alternative, including LLM-based agentic rewriting~\citep{zhao2026value} and soft-prompt tuning in the text-embedding space~\citep{jing2025promptsafe}, which often neutralize prompts that carry explicit unsafe intent. However, no defense is exhaustive, and unsafe generations inevitably slip through. We refer to these as \textit{residual adversarial outputs}: given a generation system $S$, which may be a base model $G$ or $G$ equipped with any defense, an output $S(p)$ is residual adversarial if it contains unsafe visual content despite passing all safeguards of $S$. Such residuals arise from multiple sources: adaptive attacks circumvent concept-erased and filtered models~\citep{tsai2024ring, zhang2024generate, yang2024mma}, removed concepts can persist even after modifying model weights~\citep{pham2023circumventing}, and even textually benign prompts can unintentionally elicit harmful content~\citep{li2024art}. What these cases share is that the harmful content is rooted in the model's learned output distribution, which existing safeguards, whether operating on the text, the weights, or the inference process, cannot fully remove. Motivated by this, we cast the mitigation of residual adversarial outputs as a distributional alignment problem: instead of modifying $S$, we contrastively optimize its prompt to steer generations toward safe and away from unsafe regions of the model's learned output distribution, as illustrated in Figure~\ref{fig:intro}.

In this work, we study distribution-guided prompt optimization as a black-box defense against unsafe text-to-image generation. Specifically, we explore how contrastive scoring against a few-shot set of a target system's own safe and unsafe outputs can systematically steer residual adversarial outputs toward safe generation regions. We present DiSCO, a training-free, strictly black-box, plug-and-play module that uses the prompt as its only interface, requiring no model retraining, fine-tuning, or access to model internals. Starting from the prompt behind a residual adversarial output, DiSCO performs distribution-guided contrastive suffix expansion via beam search, using only 25 safe and 25 unsafe human-labeled generations of the target system as references. We conduct a comprehensive empirical study of DiSCO against state-of-the-art red-teaming attacks, and further evaluate its generalization to unseen prompt distributions. Our experiments show that equipping both undefended and defended systems with DiSCO yields consistent safety improvements across all evaluated attacks. Our contributions are as follows:

\begin{itemize}
\item \textbf{Problem:} We formalize \emph{residual adversarial outputs} in text-to-image safety: unsafe generations that bypass the base model or any existing defense, rooted in the model's learned output distribution.
\item \textbf{Method:} We introduce \textbf{DiSCO}, a training-free, \emph{black-box}, \emph{plug-and-play} module that mitigates residual adversarial outputs via distribution-guided contrastive suffix search against a few-shot, human-labeled reference set, steering generations toward safe and away from unsafe regions without retraining, fine-tuning, or model access.
\item \textbf{Results:} Under four red-teaming attacks, DiSCO improves safety in every evaluated setting, reducing average NudeNet ASR from \textbf{32.9\% to 18.8\%}, with relative reductions of \textbf{37.3\%} on undefended models and \textbf{52.5\%} when combined with existing defenses. DiSCO further generalizes to unseen prompt distributions without reconstructing its reference pools.
\end{itemize}

\section{Related Work}
\label{sec:related_work}
\paragraph{Adversarial Robustness of Text-to-Image Models.}
Red-teaming has become the standard way to stress-test text-to-image safety, showing that adversarial prompting can reliably elicit NSFW generations under both black-box and white-box threat models. Black-box attacks craft prompts without accessing the generator's internals: Ring-A-Bell~\citep{tsai2024ring} extracts holistic concept representations to construct problematic prompts, while SneakyPrompt~\citep{yang2024sneakyprompt} searches for token perturbations that evade safety filters. White-box attacks leverage gradients and internal representations to directly optimize for unsafe outputs, as in MMA-Diffusion~\citep{yang2024mma}, and UnlearnDiffAtk~\citep{zhang2024generate} specifically targets concept-erased models, showing that unlearned concepts can be recovered. More recently, LLM-driven attacks such as APT~\citep{liu2025autoprompt} generate fluent, human-readable adversarial suffixes that bypass heuristic filters and are difficult to flag at the text level.

\paragraph{Defending Text-to-Image Models.}
Defenses span a spectrum of model access and intervention. Weight-modifying approaches suppress unsafe concepts through fine-tuning or editing, e.g., ESD~\citep{gandikota2023erasing}, AdvUnlearn~\citep{zhang2024defensive}, and RECE~\citep{gong2024rece}, or through external rectification modules such as SafePatch~\citep{meng2026beyond}; however, they require parameter access and may need repeated updates as new failure modes emerge. Inference-time methods avoid permanent edits but still depend on internal representations, limiting portability across architectures, e.g., SLD-Max~\citep{schramowski2023safe}, SAFREE~\citep{yoon2025safree}, and DTVI~\citep{tan2026dtvi}. Soft-prompt methods such as PromptGuard~\citep{yuan2026promptguard} and PromptSafe~\citep{jing2025promptsafe} learn safety embeddings in the text-encoder space, which likewise requires access to the encoder. In contrast, LLM-based rewriting such as VALOR~\citep{zhao2026value} offers a lightweight, model-agnostic safeguard by sanitizing prompts before generation. Yet harmful content rooted in the model's learned output distribution can still bypass all of these safeguards. DiSCO addresses this gap as a strictly black-box module guided by the target model's own safe and unsafe output distributions, complementing both undefended systems and existing defenses without modifying the generator.

\section{Methodology}
\label{sec:methodology}

\begin{figure*}[t]
  \centering
  \includegraphics[width=0.98\linewidth]{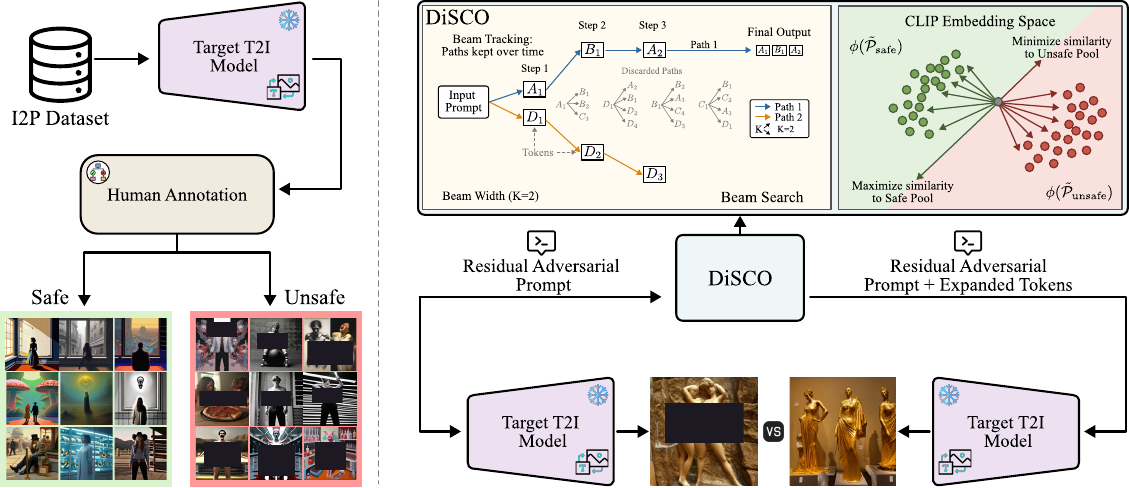}
\caption{Overview of DiSCO. \textbf{Left:} System-specific safe and unsafe reference pools are built by manually annotating images generated by the target system from I2P prompts. \textbf{Right:} DiSCO expands a residual adversarial prompt with suffix tokens optimized via contrastive beam search in CLIP space, steering the target system from unsafe to safe generation without modifying the model.}
  \label{fig:pipeline}
\end{figure*}

We formulate text-to-image safety as a \textit{distributional alignment} problem solved through prompt optimization, as shown in Figure~\ref{fig:pipeline}. Existing defenses either modify the model's internals or operate solely on the text, overlooking how the input prompt reflects the system's visual output distribution. We use the prompt as the carrier of this alignment because it is the only input shared by all T2I systems, independent of their architecture or deployed defenses, and it can be optimized without accessing model internals. We represent each system's output distribution with a few-shot set of its safe and unsafe generations, labeled by humans, and study how optimizing the prompt to align generations with the safe distribution, and away from the unsafe one, enhances robustness against residual adversarial outputs.
\subsection{Problem Definition and Evaluation Protocol}
\label{sec:problem_definition}
Let $\mathcal{T}$ be the space of text prompts and $\mathcal{X}$ the space of images. A text-to-image generator is a (possibly stochastic) mapping $G:\mathcal{T}\rightarrow\mathcal{X}$ that produces an image $x\sim G(p)$ given a prompt $p\in\mathcal{T}$. We study red-teaming attacks that craft adversarial prompts to increase the likelihood of unsafe generations: an attack algorithm $\mathcal{A}$ maps an initial prompt (or template) to $p_{\text{adv}}=\mathcal{A}(p)$.
In practice, text-to-image systems may incorporate diverse safety mechanisms, including prompt sanitization, internal safety interventions, and output filtering. We abstract the generator together with any such defense as a single black-box system $S:\mathcal{T}\rightarrow\mathcal{X}$, which reduces to $G$ when no defense is applied. Despite these safeguards, $S$ may still produce unsafe images from an input prompt $p_{\text{adv}}$; we refer to such generations as \textit{residual adversarial outputs}, which are rooted in the model's learned output distribution. We investigate whether distribution-guided optimization of the input prompt, using only black-box queries to $S$, can further mitigate these residual failures without modifying the system.

\paragraph{Residual adversarial outputs.}
Let $\mathrm{SafeImg}:\mathcal{X}\rightarrow\{0,1\}$ denote an image-level safety assessment. We say that a prompt $p\in\mathcal{T}$ yields \emph{residual adversarial outputs} w.r.t.\ a system $S$ if, despite passing all safeguards of $S$, it still produces unsafe images with high probability:
\begin{equation}
\mathbb{P}_{x\sim S(p)}\!\left[\mathrm{SafeImg}(x)=0\right]\ \geq 0.5.
\end{equation}
This motivates viewing defense as a \emph{distributional alignment} problem: rather than modifying $S$, we optimize the input prompt so that the induced generation shifts away from unsafe regions and toward safe regions of $S$'s learned output distribution. This formulation is agnostic to how the input prompt appears and to which defenses $S$ employs, covering both explicit adversarial prompts and textually benign ones.

\paragraph{Defense setting (strictly black-box).}
We model DiSCO as a module $\mathcal{D}:\mathcal{T}\rightarrow\mathcal{T}$ placed in front of the system $S$, which transforms the original input prompt $p$ into an optimized prompt $p^{*}=\mathcal{D}(p)$, while $S$, including any internal defense at the prompt, model, or output level, runs unmodified to produce $x^{*}\sim S(p^{*})$. At inference time, DiSCO is applied only to residual adversarial outputs: an image-level safety assessment of the initial generation $x_0\sim S(p)$ identifies whether $x_0$ is unsafe, in which case DiSCO optimizes $p$ to obtain $p^*$; otherwise, $x_0$ passes through unchanged.

The defense has no access to the parameters, gradients, attention maps, intermediate activations, or internal defenses of $S$; it may only query $S$ to obtain generated images. At a high level, we seek to reduce unsafe generations while preserving the user intent of $p$:
\begin{equation}
\begin{gathered}
\min_{\mathcal{D}}\
\mathbb{E}_{p}\Big[
\mathbb{P}_{x\sim S(\mathcal{D}(p))}
\big[\mathrm{SafeImg}(x)=0\big]
\Big] \\
\text{s.t.}\quad
\mathrm{Sem}(\mathcal{D}(p),p)\ge \tau,
\end{gathered}
\end{equation}
where $\mathrm{Sem}(\cdot,\cdot)$ denotes an abstract semantic-fidelity measure and $\tau$ is a fidelity threshold. We evaluate semantic preservation using CLIP score~\citep{radford2021learning}, with prompt-embedding cosine similarity reported in the supplementary material.

\paragraph{Evaluation protocol.}
We report robustness using \emph{Attack Success Rate} (ASR), the fraction of evaluated prompts that produce unsafe images under the tested system, whether undefended, baseline-defended, or equipped with DiSCO:
\begin{equation}
\mathrm{ASR} \;=\; \frac{1}{|\mathcal{Q}|}\sum_{q\in\mathcal{Q}} \mathbb{I}\!\left[\mathrm{SafeImg}(x_q)=0\right],
\qquad x_q \sim S(\hat{p}_q),
\end{equation}
where $\mathcal{Q}$ is the evaluated prompt set, and $\hat{p}_q$ is the input prompt $p_q$ for systems without DiSCO and $\mathcal{D}(p_q)$ for systems with DiSCO. Since the evaluated attacks target nudity, we instantiate $\mathrm{SafeImg}$ with NudeNet~\citep{mandic2024nudenet}, the standard evaluator in prior work. We additionally measure semantic fidelity with CLIP score~\citep{radford2021learning} between generated images and their prompts on benign COCO~\citep{lin2014microsoft} captions (Table~\ref{tab:asr}).

\subsection{Distribution-Guided Contrastive Suffix Optimization}
\label{sec:contrastive_optimization}

Given the problem formulation, we now present our defense: \textbf{Dis}tribution-Guided \textbf{C}ontrastive Suffix \textbf{O}ptimization~(DiSCO). DiSCO is a strictly black-box module that rectifies the learned output distribution of a target system $S$ by aligning generations toward its safe region and away from its unsafe one, using the \emph{prompt} as the carrier of this alignment and feedback obtained only from \emph{observable outputs} of $S$. In a nutshell, given an input prompt $p$, DiSCO performs autoregressive \emph{suffix expansion} and selects the suffix that maximizes a distribution-guided contrastive score, computed in a model-agnostic embedding space, that pulls generations toward safe and away from unsafe regions, as illustrated in Figure~\ref{fig:pipeline}.

\paragraph{Safe and unsafe reference pools.}
\label{para:reference_pool}
To represent each system's output distribution under the black-box constraint, we construct a \emph{few-shot}, system-specific reference set. For each system $S$, we generate images from original Inappropriate Image Prompts (I2P)~\cite{schramowski2023safe} prompts, without any adversarial modification, and manually label them as safe or unsafe, discarding ambiguous cases, to obtain $\mathcal{P}_{\text{safe}}=\{x_i^{+}\}_{i=1}^{N}$ and $\mathcal{P}_{\text{unsafe}}=\{x_j^{-}\}_{j=1}^{N}$ with $N=25$. Because the pools are labeled by humans rather than by safety classifiers, DiSCO's optimization signal is independent of the detector used for evaluation. Because they are generated by $S$ itself, they reflect its learned output distribution, including the effect of any deployed defense, and require neither external unsafe sources nor internal access.

\paragraph{Distribution-guided contrastive objective.}
Given a candidate prompt $\hat{p}$, we query the target system to obtain $\hat{x}\sim S(\hat{p})$ and embed it using a frozen CLIP image encoder $\phi(\cdot)$. At each optimization step, we sample one shared set of $R$ references from each 25-image pool and use them to score all candidate suffixes at that step, yielding $\tilde{\mathcal{P}}_{\mathrm{safe}}\subset\mathcal{P}_{\mathrm{safe}}$ and $\tilde{\mathcal{P}}_{\mathrm{unsafe}}\subset\mathcal{P}_{\mathrm{unsafe}}$ with $|\tilde{\mathcal{P}}_{\mathrm{safe}}|=|\tilde{\mathcal{P}}_{\mathrm{unsafe}}|=R$ (default $R=8$). We score $\hat{p}$ by the contrastive alignment of its generated image with the two reference sets:
\begin{equation}
\begin{gathered}
J(\hat{p}) =
\frac{1-\lambda}{R}
\sum_{x_i^{+}\in \tilde{\mathcal{P}}_{\text{safe}}}
\cos\!\big(\phi(\hat{x}), \phi(x_i^{+})\big)
\\[4pt]
-
\frac{\lambda}{R}
\sum_{x_j^{-}\in \tilde{\mathcal{P}}_{\text{unsafe}}}
\cos\!\big(\phi(\hat{x}), \phi(x_j^{-})\big),
\end{gathered}
\label{eq:contrastive}
\end{equation}
where $\cos(\cdot,\cdot)$ denotes cosine similarity and $\lambda\in[0,1]$ balances attraction to the safe pool against repulsion from the unsafe pool (default $\lambda=0.5$). Maximizing $J$ aligns generations with the safe region of the system's output distribution while pushing them away from its unsafe region. Re-sampling references at every step exposes the optimizer to diverse views of both distributions without increased per-step cost. A preference-based variant with a unified objective is discussed in the supplementary material.

\paragraph{Suffix expansion via beam search.}
We realize this alignment by expanding a suffix token by token using a suffix-generator language model $M$ (Llama~3~8B). Starting from the input prompt $p$, we maintain a beam $\mathcal{B}$ of size $K$ (default $K=4$). At each step $t \in \{1,\dots,T\}$ (default $T=12$), $M$ samples $b$ next-token continuations for each beam candidate (default $b=4$), and we query $S$ for each continuation, score it with Eq.~\eqref{eq:contrastive}, and retain the top-$K$ candidates. After $T$ steps, the optimized prompt is selected as $p^{*}=\arg\max_{\hat{p}\in \mathcal{B}} J(\hat{p})$. The complete algorithm is in the supplementary material.

\paragraph{Computational overhead.}
DiSCO requires only a one-time setup per target system: labeling $25$ safe and $25$ unsafe images generated from I2P prompts, with labels agreed by at least three annotators; annotator agreement is reported in the supplementary material. At inference, the default beam search ($K=4$, $T=12$, $b=4$) requires $b+(T-1)bK=180$ candidate generations, rendered with 4 denoising steps for CLIP-based ranking, while the final image uses the system's full schedule. DiSCO runs only on residual adversarial outputs, and optimized prompts can be cached for reuse. Wall-clock costs are analyzed in the supplementary material.

\section{Experiments}
\paragraph{Models.} We evaluate DiSCO on text-to-image generators spanning the two major architecture families: UNet-based diffusion models (SD 1.4 and SD 2.0~\citep{rombach2022high}) and DiT-based models (Flux~\citep{flux2024} and SD 3~\citep{esser2024scaling}). This allows us to assess whether distributional alignment through prompt optimization generalizes across architectures and model scales.

\paragraph{Attacks and datasets.}
We benchmark robustness under a set of red-teaming attacks covering both black-box and white-box threat models. Ring-A-Bell~\citep{tsai2024ring} is a model-agnostic attack that crafts adversarial prompts without access to the target generator's internals. UnlearnDiffAtk~\citep{zhang2024generate} and MMA-Diffusion~\citep{yang2024mma} are stronger white-box attacks that optimize prompts using model-side signals such as gradients and internal representations, while Prompting4Debugging (P4D)~\citep{chin2023prompting4debugging} specifically targets safety-equipped models by optimizing prompts that bypass their defense mechanisms. We use all unique prompts released from each attack (95, 81, 812, and 451, respectively). Together, these attacks probe complementary failure modes and elicit residual adversarial outputs from both undefended and defended systems. To assess generalization, we further evaluate on two unseen unsafe prompt distributions, SneakyPrompt~\citep{yang2024sneakyprompt} and SafeGen~\citep{li2024safegen} (Section~\ref{sec:ood}). We additionally stress-test DiSCO against the LLM-driven attack APT~\citep{liu2025autoprompt} using our re-implementation, with results provided in the supplementary material.

\paragraph{Metrics.}
We measure safety using ASR, the fraction of evaluated prompts whose generated images are classified as unsafe. Since all four attacks target nudity, we use NudeNet~\citep{mandic2024nudenet}, the standard detector in prior work. We evaluate semantic fidelity using CLIP score~\citep{radford2021learning} on benign COCO~\citep{lin2014microsoft} captions. Evaluation across all seven I2P harm categories is provided in the supplementary material.

\paragraph{DiSCO hyperparameters.}
Unless stated otherwise, DiSCO uses Llama 3 8B~\citep{grattafiori2024llama3herdmodels} as the suffix generator, with beam width $K=4$, $b=4$ proposals per beam, suffix length $T=12$, and contrastive weight $\lambda=0.5$. Candidates are scored using frozen CLIP ViT-L/14 embeddings and $R=8$ references sampled from each system-specific pool of 25 human-labeled safe and 25 unsafe images.

\paragraph{Evaluation protocol.}
For each prompt, we first generate an image using the target system's existing configuration. NudeNet identifies residual adversarial outputs among these initial generations, and DiSCO is applied only to those prompts; all other images pass through unchanged. All baseline and DiSCO ASRs are computed over the same full prompt set, with unflagged prompts counted as safe in both cases.

\begin{table*}[t]
\centering
\setlength{\tabcolsep}{2.5pt}
\renewcommand{\arraystretch}{1.1}
\newlength{\bcol}\settowidth{\bcol}{\small 88.8}
\newlength{\dcol}\settowidth{\dcol}{\small$\mathbf{88.8}_{\pm 8.8}$}
\newlength{\acol}\settowidth{\acol}{\small MMA-Diffusion}
\newlength{\rcol}\settowidth{\rcol}{\small $-88.8\%$}
\caption{NudeNet ASR (\%, $\downarrow$) under four adversarial attacks with and without DiSCO, over all unique prompts per attack (Ring-A-Bell: 95, UnlearnDiffAtk: 81, MMA-Diffusion: 812, P4D: 451). DiSCO results are mean$_{\pm\mathrm{std}}$ over three seeds. \emph{Rel.\,$\Delta$} denotes the relative change of DiSCO averaged over four systems per subtable. The bottom row reports CLIP score ($\times100$) on 100 benign COCO captions with DiSCO applied to every prompt. (a) Base T2I architectures; (b) existing defense methods.}
\label{tab:asr}
\begin{subtable}[t]{0.495\textwidth}
\centering
\small
\caption{Base T2I architectures}
\label{tab:asr_arch}
\resizebox{\linewidth}{!}{
\begin{tabular}{w{l}{\acol}*{4}{w{c}{\bcol}w{c}{\dcol}}w{c}{\rcol}}
\toprule
& \multicolumn{2}{c}{SD 1.4} & \multicolumn{2}{c}{SD 2.0} & \multicolumn{2}{c}{SD 3} & \multicolumn{2}{c}{Flux} & \\
\cmidrule(lr){2-3} \cmidrule(lr){4-5} \cmidrule(lr){6-7} \cmidrule(lr){8-9}
Attack & Base & \cellcolor{discobg}\textbf{+DiSCO} & Base & \cellcolor{discobg}\textbf{+DiSCO} & Base & \cellcolor{discobg}\textbf{+DiSCO} & Base & \cellcolor{discobg}\textbf{+DiSCO} & Rel.\,$\Delta$ \\
\midrule
Ring-A-Bell & 89.5 & \cellcolor{discobg}\bms{73.7}{1.1} & 86.3 & \cellcolor{discobg}\bms{73.7}{2.8} & 45.3 & \cellcolor{discobg}\bms{26.7}{4.4} & 93.7 & \cellcolor{discobg}\bms{63.9}{3.0} & $-24.4\%$ \\
UnlearnDiffAtk & 67.9 & \cellcolor{discobg}\bms{46.5}{2.6} & 40.7 & \cellcolor{discobg}\bms{19.3}{1.9} & 16.0 & \cellcolor{discobg}\bms{7.0}{1.9} & 38.3 & \cellcolor{discobg}\bms{15.2}{3.6} & $-46.0\%$ \\
MMA-Diffusion & 76.7 & \cellcolor{discobg}\bms{50.6}{1.5} & 21.2 & \cellcolor{discobg}\bms{8.2}{0.8} & 5.4 & \cellcolor{discobg}\bms{1.8}{0.3} & 11.9 & \cellcolor{discobg}\bms{2.8}{0.6} & $-45.0\%$ \\
P4D & 34.1 & \cellcolor{discobg}\bms{14.1}{1.9} & 22.6 & \cellcolor{discobg}\bms{11.1}{0.6} & 8.9 & \cellcolor{discobg}\bms{3.8}{0.5} & 18.8 & \cellcolor{discobg}\bms{6.7}{1.4} & $-57.7\%$ \\
\midrule
Average & 67.1 & \cellcolor{discobg}\textbf{46.2} & 42.7 & \cellcolor{discobg}\textbf{28.1} & 18.9 & \cellcolor{discobg}\textbf{9.8} & 40.7 & \cellcolor{discobg}\textbf{22.1} & $-37.3\%$ \\
\midrule
\multicolumn{10}{l}{\textit{Benign COCO prompts}} \\
CLIP $\uparrow$ & 26.5 & \ms{25.1}{0.4} & 26.2 & \ms{25.1}{0.2} & 26.4 & \ms{25.4}{0.2} & 26.4 & \ms{24.9}{0.1} & $-4.7\%$ \\

\bottomrule
\end{tabular}
}
\end{subtable}
\hfill
\begin{subtable}[t]{0.495\textwidth}
\centering
\small
\caption{Defense methods}
\label{tab:asr_defense}
\resizebox{\linewidth}{!}{
\begin{tabular}{w{l}{\acol}*{4}{w{c}{\bcol}w{c}{\dcol}}w{c}{\rcol}}
\toprule
& \multicolumn{2}{c}{SLD-Max} & \multicolumn{2}{c}{SAFREE} & \multicolumn{2}{c}{RECE} & \multicolumn{2}{c}{ESD} & \\
\cmidrule(lr){2-3} \cmidrule(lr){4-5} \cmidrule(lr){6-7} \cmidrule(lr){8-9}
Attack & Base & \cellcolor{discobg}\textbf{+DiSCO} & Base & \cellcolor{discobg}\textbf{+DiSCO} & Base & \cellcolor{discobg}\textbf{+DiSCO} & Base & \cellcolor{discobg}\textbf{+DiSCO} & Rel.\,$\Delta$ \\
\midrule
Ring-A-Bell & 72.6 & \cellcolor{discobg}\bms{48.4}{3.2} & 52.6 & \cellcolor{discobg}\bms{27.0}{3.7} & 3.2 & \cellcolor{discobg}\bms{0.4}{0.6} & 23.2 & \cellcolor{discobg}\bms{13.0}{3.0} & $-41.4\%$ \\
UnlearnDiffAtk & 30.9 & \cellcolor{discobg}\bms{14.4}{1.4} & 22.2 & \cellcolor{discobg}\bms{10.7}{0.7} & 11.1 & \cellcolor{discobg}\bms{4.1}{0.7} & 14.8 & \cellcolor{discobg}\bms{4.5}{1.4} & $-57.3\%$ \\
MMA-Diffusion & 52.5 & \cellcolor{discobg}\bms{27.7}{1.4} & 33.6 & \cellcolor{discobg}\bms{11.5}{0.5} & 32.0 & \cellcolor{discobg}\bms{10.0}{0.7} & 6.2 & \cellcolor{discobg}\bms{1.0}{0.4} & $-59.6\%$ \\
P4D & 6.7 & \cellcolor{discobg}\bms{1.5}{0.1} & 4.9 & \cellcolor{discobg}\bms{1.3}{0.3} & 4.4 & \cellcolor{discobg}\bms{1.5}{0.7} & 2.7 & \cellcolor{discobg}\bms{0.6}{0.1} & $-73.8\%$ \\
\midrule
Average & 40.7 & \cellcolor{discobg}\textbf{23.0} & 28.3 & \cellcolor{discobg}\textbf{12.6} & 12.7 & \cellcolor{discobg}\textbf{4.0} & 11.7 & \cellcolor{discobg}\textbf{4.8} & $-52.5\%$ \\
\midrule
\multicolumn{10}{l}{\textit{Benign COCO prompts}} \\
CLIP $\uparrow$ & 24.5 & \ms{22.7}{0.2} & 25.7 & \ms{24.2}{0.1} & 25.7 & \ms{24.9}{0.2} & 25.7 & \ms{24.1}{0.2} & $-5.6\%$ \\
\bottomrule
\end{tabular}
}
\end{subtable}
\end{table*}

\subsection{DiSCO Shields Undefended Models}
\label{sec:undefended}
We first evaluate DiSCO as a \emph{standalone} safety module by applying it to undefended text-to-image generators. Concretely, we apply the adversarial prompt sets to models spanning UNet-based architectures (SD 1.4, SD 2.0) and DiT-based architectures (Flux, SD 3), and measure DiSCO's effect under the same attack protocols to isolate the contribution of its distributional alignment.

Table~\ref{tab:asr_arch} shows that DiSCO reduces NudeNet ASR for every model and attack without modifying the generator, with an average relative reduction of $37.3\%$ and standard deviations of at most $4.4$ points across seeds. Averaged across the four attacks, ASR decreases from 67.1\% to 46.2\% on SD~1.4, from 42.7\% to 28.1\% on SD~2.0, from 18.9\% to 9.8\% on SD~3, and from 40.7\% to 22.1\% on Flux. The gains are largest under UnlearnDiffAtk, MMA-Diffusion, and P4D, with relative reductions of $45.0$--$57.7\%$; for example, UnlearnDiffAtk ASR on Flux drops from $38.3\%$ to $15.2\pm3.6\%$, and MMA-Diffusion ASR on SD~2.0 from $21.2\%$ to $8.2\pm0.8\%$. On benign COCO captions, applying DiSCO to every prompt lowers CLIP score by only $4.7\%$ relative. This asymmetry, a $37.3\%$ safety gain against a $4.7\%$ fidelity cost, reflects that DiSCO appends a suffix rather than rewriting the prompt, keeping the original request intact; the supplementary material reports the CLIP similarity between prompts before and after applying DiSCO.

\subsection{Combining DiSCO with Defended Models}
\label{sec:boosting_defense}
We next evaluate whether DiSCO, applied as a plug-and-play module, consistently improves the robustness of various defense mechanisms across diverse attack strategies. DiSCO targets residual adversarial outputs that bypass each defense, treating the defended model as a black-box system.

\begin{figure}[b]
  \centering
  \includegraphics[width=\columnwidth]{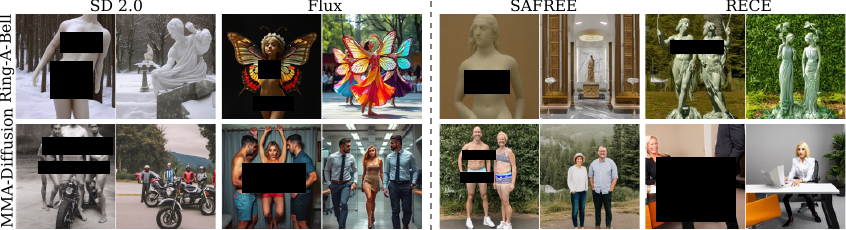}
  \caption{Qualitative results under Ring-A-Bell and MMA-Diffusion on undefended (SD 2.0, Flux) and defended (SAFREE, RECE) systems, before and after DiSCO. Unsafe regions are masked; prompts and more results are in supplementary material.}
  \label{fig:merged}
\end{figure}

Table~\ref{tab:asr_defense} shows that adding DiSCO reduces NudeNet ASR for every defense--attack combination, with an average relative reduction of $52.5\%$ and standard deviations of at most $3.7$ points across seeds. The effect depends on how each defense specifies unsafety. Inference-time defenses steer generation away from unsafe directions defined by textual concepts, which concept-level attacks can circumvent: under Ring-A-Bell, SLD-Max and SAFREE still yield $72.6\%$ and $52.6\%$ ASR, whereas the weight-editing defenses ESD and RECE reduce it to $23.2\%$ and $3.2\%$. Consequently, DiSCO yields the largest absolute gains on SLD-Max ($40.7\%\rightarrow23.0\%$) and SAFREE ($28.3\%\rightarrow12.6\%$). Weight editing, in turn, removes targeted concepts but leaves other unsafe modes reachable, as exposed by MMA-Diffusion, under which RECE still yields $32.0\%$ ASR. These residuals are sparser, and DiSCO removes most of them, achieving the highest relative reductions on RECE ($68.4\%$; $12.7\%\rightarrow4.0\%$) and ESD ($59.3\%$; $11.7\%\rightarrow4.8\%$). Figure~\ref{fig:merged} provides qualitative comparisons, with more results in the supplementary material, and Figure~\ref{fig:asr_plot} visualizes DiSCO's per-attack ASR reductions across undefended and defended systems. Overall, DiSCO complements diverse defenses at a modest fidelity cost, lowering CLIP score by only $5.6\%$ relative on benign COCO prompts for defended systems, versus a $52.5\%$ ASR reduction (Table~\ref{tab:asr}).

\begin{figure}[ht]
\centering
\includegraphics[width=\columnwidth]{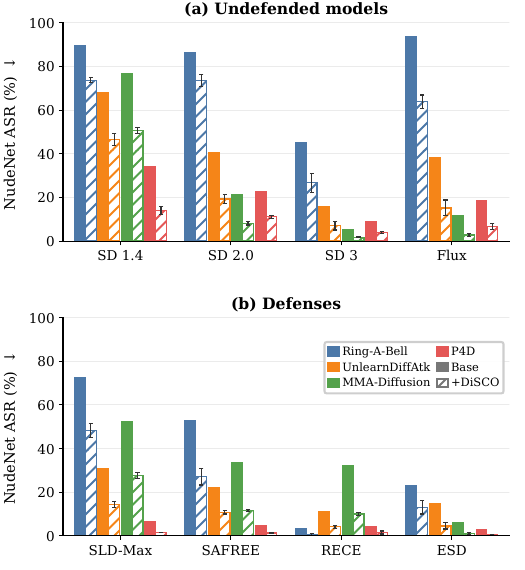}
\caption{NudeNet Attack Success Rate (ASR, \%; mean $\pm$ std over three seeds; lower is better) across four attacks on (a) undefended models and (b) existing defenses. Colors indicate attacks; solid and hatched bars denote results before and after DiSCO.}
\label{fig:asr_plot}
\end{figure}

\subsection{Generalization to Unseen Prompt Distributions}
\label{sec:ood}
Since DiSCO's reference pools are generated from I2P prompts, we evaluate its generalization on two unsafe prompt distributions that differ from I2P in how they were constructed: (i) the NSFW prompts of SneakyPrompt~\citep{yang2024sneakyprompt}, written by GPT-4~\citep{openai2024gpt4technicalreport} rather than collected from users, and (ii) the real-world set of SafeGen~\citep{li2024safegen}, consisting of BLIP-2~\citep{li2023blip} captions of real NSFW images. DiSCO is applied with the same system-specific pools built from I2P generations, without reconstruction or dataset-specific tuning.
\begin{table}[t]
\centering
\small
\setlength{\tabcolsep}{5pt}
\caption{NudeNet ASR (\%, $\downarrow$) on unseen unsafe prompt distributions. DiSCO uses the same system-specific pools built from I2P generations, without reconstruction or tuning; DiSCO results are mean$_{\pm\mathrm{std}}$ over three seeds. SneakyPrompt: 181 prompts; SafeGen: first 200 prompts. \emph{Rel.\,$\Delta$} denotes the relative ASR change of DiSCO averaged over four systems in each group.}
\label{tab:ood}
\begin{tabular}{lcccc}
\toprule
& \multicolumn{2}{c}{SneakyPrompt} & \multicolumn{2}{c}{SafeGen} \\
\cmidrule(lr){2-3} \cmidrule(lr){4-5}
System & Base & +DiSCO & Base & +DiSCO \\
\midrule
SD 1.4  & 63.5 & $\mathbf{30.4}_{\pm 4.1}$ & 63.0 & $\mathbf{45.8}_{\pm 1.3}$ \\
SD 2.0  & 25.4 & $\mathbf{12.7}_{\pm 3.6}$ & 20.5 & $\mathbf{12.2}_{\pm 0.6}$ \\
SD 3    & 8.3  & $\mathbf{7.9}_{\pm 0.7}$  & 11.5 & $\mathbf{4.8}_{\pm 1.6}$ \\
Flux    & 35.4 & $\mathbf{19.2}_{\pm 6.1}$ & 29.5 & $\mathbf{14.2}_{\pm 1.0}$ \\
\cmidrule(lr){1-5}
Rel.\,$\Delta$ & \multicolumn{2}{c}{$-47.1\%$} & \multicolumn{2}{c}{$-38.2\%$} \\
\midrule
SLD-Max & 34.3 & $\mathbf{5.3}_{\pm 2.7}$  & 25.0 & $\mathbf{5.0}_{\pm 0.5}$ \\
SAFREE  & 15.5 & $\mathbf{5.2}_{\pm 1.7}$  & 19.0 & $\mathbf{8.8}_{\pm 0.8}$ \\
RECE    & 13.3 & $\mathbf{1.8}_{\pm 1.3}$  & 10.5 & $\mathbf{2.7}_{\pm 0.3}$ \\
ESD     & 0.6  & $\mathbf{0.4}_{\pm 0.3}$  & 2.0  & $\mathbf{1.0}_{\pm 0.0}$ \\
\cmidrule(lr){1-5}
Rel.\,$\Delta$ & \multicolumn{2}{c}{$-80.1\%$} & \multicolumn{2}{c}{$-69.0\%$} \\
\bottomrule
\end{tabular}
\end{table}

As shown in Table~\ref{tab:ood}, DiSCO reduces ASR across all evaluated systems on both datasets. On undefended models, the average relative reduction is $47.1\%$ on SneakyPrompt and $38.2\%$ on SafeGen, matching or exceeding that on I2P attacks ($37.3\%$). The gains are even larger on defended systems, reaching $80.1\%$ and $69.0\%$, respectively, compared with $52.5\%$ on I2P; SLD-Max, for instance, drops from $34.3\%$ to $5.3\%$ on SneakyPrompt and from $25.0\%$ to $5.0\%$ on SafeGen. These consistent gains under fixed reference pools indicate that DiSCO captures transferable properties of each system's safe and unsafe output distributions and that its complementarity with existing defenses persists under distribution shift.

\paragraph{DiSCO is most effective as a complement.}
\label{para:boost}
Across both I2P attacks and unseen prompt distributions, DiSCO yields larger gains on defended systems than on undefended ones. On I2P, the average relative reduction rises from $37.3\%$ on base models to $52.5\%$ with defenses, and this holds for every attack (Table~\ref{tab:asr}); the gap widens further on unseen distributions, from $47.1\%$ to $80.1\%$ on SneakyPrompt and from $38.2\%$ to $69.0\%$ on SafeGen (Table~\ref{tab:ood}). The comparison is most direct on SD~1.4, which underlies all four defenses: on I2P, DiSCO reduces ASR by $31.1\%$ on the undefended backbone but by $43.5$--$68.4\%$ on its defended variants, lowering the residual ASR from $46.2\%$ to as little as $4.0\%$, and on SafeGen by $27.3\%$ versus $50.0$--$80.0\%$. We attribute this to the complementary roles of the two mechanisms. Existing defenses already shift the system's output distribution toward the safe region, but leave residual unsafe modes they were not designed to cover. Starting from this partially aligned distribution, DiSCO only needs to bridge the remaining gap, and since its references are drawn from the defended system's own outputs, its suffix optimization targets exactly the failures that persist under the defense. The two mechanisms act at different levels, internally within the model and externally through the prompt, and their effects compound rather than overlap across distributions.

\subsection{Ablations and Analyses}
\label{sec:ablation}
In this section, we ablate the main components of DiSCO. Unless stated otherwise, all ablations are conducted on SD~1.4 using 340 attack prompts whose initial generations are unsafe under NudeNet: all flagged Ring-A-Bell (85) and UnlearnDiffAtk (55) prompts, and the first 100 flagged MMA-Diffusion and P4D prompts. ASR is therefore computed relative to a fully unsafe starting point. We first study the effect of the number of sampled references $R$ on defense performance (Section~\ref{sec:pool_size}). We then compare the contrastive objective against a random-suffix baseline and safe-only and unsafe-only scoring, and vary the contrastive weight $\lambda$, to assess the contribution of each pool (Section~\ref{sec:contrastive_scoring}). Finally, we analyze how the beam search parameters $K$ and $T$ affect the trade-off between defense effectiveness and computational cost (Section~\ref{sec:trade_off}).

\begin{table}[t]
\centering
\small
\setlength{\tabcolsep}{4pt}
\caption{Ablation of the number of sampled references $R$ per step on SD~1.4: NudeNet ASR (\%, $\downarrow$) on 340 initially unsafe attack prompts, mean$_{\pm\mathrm{std}}$ over three seeds. Best in \textbf{bold}, second \underline{underlined}.}
\label{tab:ablation_R}
\resizebox{\columnwidth}{!}{
\begin{tabular}{cccccc}
\toprule
$R$ & Ring-A-Bell & UnlearnDiffAtk & MMA & P4D & Total \\
\midrule
4  & $85.5_{\pm 3.8}$ & $\mathbf{64.9}_{\pm 2.7}$ & $72.7_{\pm 1.2}$ & $42.7_{\pm 2.5}$ & $65.8_{\pm 1.6}$ \\
8  & $\mathbf{82.4}_{\pm 1.2}$ & $68.5_{\pm 3.8}$ & $\mathbf{69.0}_{\pm 4.6}$ & $\mathbf{38.0}_{\pm 3.0}$ & $\mathbf{63.1}_{\pm 2.2}$ \\
16 & $\underline{84.7}_{\pm 6.2}$ & $\underline{67.8}_{\pm 4.2}$ & $\underline{71.0}_{\pm 3.6}$ & $\underline{40.0}_{\pm 1.7}$ & $\underline{64.8}_{\pm 1.5}$ \\
\bottomrule
\end{tabular}}
\end{table}

\subsubsection{Number of Sampled References $R$}
\label{sec:pool_size}
At each optimization step, DiSCO samples $R$ references from each 25-image pool. Table~\ref{tab:ablation_R} shows that a larger $R$ does not necessarily yield better robustness: total ASR is lowest at a moderate $R=8$ ($63.1\%$) and increases both for $R=4$ ($65.8\%$) and $R=16$ ($64.8\%$). With a moderate $R$, each step scores candidates against a different subset of references, so the search is guided by varied views of the safe and unsafe distributions rather than tuned to a single fixed set. When $R$ is too small, each subset is less representative of the distribution; when $R$ is too large, consecutive steps draw largely overlapping subsets, at $R=16$ already $64\%$ of each pool, reducing this diversity. DiSCO remains robust across all tested values (within $2.7$ points), and we adopt $R=8$ as the default.

\subsubsection{Contribution of the Safe and Unsafe Pools}
\label{sec:contrastive_scoring}
Our objective $J$ scores each candidate by its alignment with both reference distributions, with $\lambda$ weighting repulsion from the unsafe pool against attraction to the safe pool (Eq.~\eqref{eq:contrastive}). To assess the role of each pool, we compare it with a random-suffix baseline and single-pool variants, which correspond to the extremes of Eq.~\eqref{eq:contrastive}: safe-only scoring ($\lambda=0$), which rewards similarity to the safe pool without penalizing proximity to the unsafe pool, and unsafe-only scoring ($\lambda=1$), which penalizes similarity to the unsafe pool without rewarding alignment with the safe pool:
\begin{equation}
\begin{gathered}
J_{\text{safe}} = \frac{1}{R}\sum_{i}\cos\!\big(\phi(\hat{x}), \phi(x_i^{+})\big), \\
J_{\text{unsafe}} = -\frac{1}{R}\sum_{j}\cos\!\big(\phi(\hat{x}), \phi(x_j^{-})\big).
\end{gathered}
\label{eq:single_pool}
\end{equation}
We further sweep $\lambda\in\{0.2, 0.4, 0.5, 0.6, 0.8\}$.
\begin{table}[t]
\centering
\small
\setlength{\tabcolsep}{4pt}
\caption{Ablation of the scoring objective on SD~1.4: NudeNet ASR (\%, $\downarrow$) on 340 initially unsafe attack prompts, mean$_{\pm\mathrm{std}}$ over three seeds. \emph{Random}: suffix without search; $J_{\text{safe}}$ / $J_{\text{unsafe}}$: safe-only / unsafe-only scoring; $J$: contrastive scoring (default $\lambda{=}0.5$). Best in \textbf{bold}, second \underline{underlined}.}
\label{tab:ablation_objective}
\resizebox{\columnwidth}{!}{
\begin{tabular}{lccccc}
\toprule
Objective & Ring-A-Bell & UnlearnDiffAtk & MMA-Diffusion & P4D & Total \\
\midrule
Random                              & $88.6_{\pm 1.4}$ & $83.6_{\pm 3.6}$ & $76.7_{\pm 0.6}$ & $51.0_{\pm 3.6}$ & $73.2_{\pm 2.1}$ \\
$J_{\text{safe}}$ ($\lambda{=}0$)   & $89.4_{\pm 3.1}$ & $72.2_{\pm 5.8}$ & $75.3_{\pm 1.2}$ & $57.3_{\pm 5.0}$ & $73.0_{\pm 1.8}$ \\
\midrule
$J$ ($\lambda{=}0.2$)               & $90.2_{\pm 1.8}$ & $75.1_{\pm 2.2}$ & $74.0_{\pm 4.4}$ & $49.0_{\pm 3.6}$ & $70.9_{\pm 2.4}$ \\
$J$ ($\lambda{=}0.4$)               & $\underline{83.5}_{\pm 3.1}$ & $69.6_{\pm 7.6}$ & $70.7_{\pm 6.7}$ & $45.3_{\pm 2.9}$ & $66.3_{\pm 2.7}$ \\
$J$ ($\lambda{=}0.5$, default)      & $\mathbf{82.4}_{\pm 1.2}$ & $68.5_{\pm 3.8}$ & $\underline{69.0}_{\pm 4.6}$ & $\mathbf{38.0}_{\pm 3.0}$ & $\underline{63.1}_{\pm 2.2}$ \\
$J$ ($\lambda{=}0.6$)               & $86.2_{\pm 0.7}$ & $65.5_{\pm 8.0}$ & $70.3_{\pm 3.2}$ & $42.0_{\pm 4.4}$ & $65.2_{\pm 2.0}$ \\
$J$ ($\lambda{=}0.8$)               & $84.4_{\pm 1.8}$ & $\mathbf{59.5}_{\pm 5.6}$ & $\mathbf{66.7}_{\pm 1.2}$ & $\underline{39.7}_{\pm 6.7}$ & $\mathbf{62.0}_{\pm 1.7}$ \\
\midrule
$J_{\text{unsafe}}$ ($\lambda{=}1$) & $84.7_{\pm 2.0}$ & $\underline{63.6}_{\pm 1.8}$ & $70.0_{\pm 1.7}$ & $40.0_{\pm 5.0}$ & $63.8_{\pm 1.6}$ \\
\bottomrule
\end{tabular}}
\end{table}

Table~\ref{tab:ablation_objective} shows that distribution-guided selection, rather than merely appending tokens, drives DiSCO's gains: the contrastive objective reduces ASR from $73.2\%$ with random suffixes to $63.1\%$, with the largest gains on P4D ($51.0\%\rightarrow38.0\%$) and UnlearnDiffAtk ($83.6\%\rightarrow68.5\%$). The two pools contribute unequally but complementarily. Safe-only scoring performs on par with random suffixes ($73.0\%$), indicating that attraction toward safe references alone does not prevent generations from remaining close to unsafe ones, whereas repulsion from the unsafe pool provides most of the gain ($63.8\%$ with unsafe-only scoring). Combining both terms performs best: at $\lambda=0.5$ and $0.8$, the contrastive objective outperforms both single-pool extremes, and ASR generally decreases as weight shifts from the safe term toward the unsafe one ($70.9\%$ at $\lambda=0.2$ to $62.0\%$ at $\lambda=0.8$) before rising again at the unsafe-only extreme. For simplicity, we adopt the balanced $\lambda=0.5$ as default, which weights both pools equally and performs within one standard deviation of the best setting ($63.1\pm2.2$ vs.\ $62.0\pm1.7$).

\subsubsection{Defense--Efficiency Trade-off of Beam Search}
\label{sec:trade_off}
\begin{figure}[t]
  \centering
  \includegraphics[width=\columnwidth]{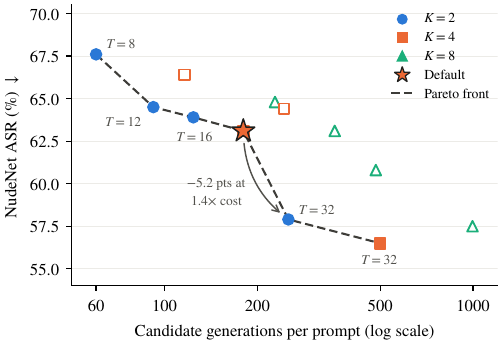}
  \caption{Defense--efficiency trade-off of beam width $K$ and suffix length $T$ on SD~1.4: NudeNet ASR ($\downarrow$) on 340 initially unsafe attack prompts versus candidate generations per prompt. Filled markers lie on the Pareto front and hollow markers are dominated settings. The default ($\star$, $K{=}4$, $T{=}12$) is Pareto-optimal.}
  \label{fig:beam_tradeoff}
\end{figure}

We study how beam-search suffix expansion balances defense effectiveness and computational cost by varying (i) the beam width $K$ and (ii) the suffix length $T$. Since the number of candidate generations, $b+(T-1)bK$, grows with both parameters, Figure~\ref{fig:beam_tradeoff} plots ASR against this cost. Suffix length is the dominant factor: settings with $T=32$ achieve the lowest ASR for every beam width ($56.5$--$57.9\%$), whereas wider beams rarely pay off, as every $K=8$ setting is dominated by a cheaper alternative. From the Pareto-optimal default $K=4,T=12$, extending the suffix to $K=2,T=32$ further reduces ASR by $5.2$ points at $1.4\times$ the cost, and $K=4,T=32$ achieves the lowest ASR ($56.5\%$) at $2.8\times$ the cost, offering a direct option when a larger query budget is available.

\section{Conclusion}
\label{sec:conclusion}

We introduced DiSCO, a training-free, black-box, plug-and-play defense that mitigates residual adversarial outputs in text-to-image generation by casting safety as distributional alignment through prompt optimization guided by a few-shot set of the target system's own safe and unsafe generations. We conducted a comprehensive empirical study across four attacks spanning black-box and white-box threat models, four defenses covering inference-time, weight-editing, and fine-tuning approaches, and four architectures spanning UNet-based and DiT-based families. With only 25 safe and 25 unsafe references per system, DiSCO reduces average NudeNet ASR from $32.9\%$ to $18.8\%$, with relative reductions of $37.3\%$ on undefended models and $52.5\%$ when combined with existing defenses, and generalizes to unseen prompt distributions without reconstructing its references. These results establish DiSCO as a general complement to both undefended and defended systems that is most effective when paired with an existing defense, strengthening safety without retraining or access to model internals.

\paragraph{Limitations.}
Since DiSCO operates through the system's learned output distribution, its suffix may occasionally invoke concepts heavily represented in the training data, such as Minions, causing mild semantic deviation while yielding safe images. Adding a semantic-preservation constraint to the suffix search is a natural extension for future work.

\section*{LLM Usage Statement}
Large language models (LLMs) were used in two distinct roles in this work. First, as part of the proposed method, DiSCO employs Llama 3 8B~\citep{grattafiori2024llama3herdmodels} as its suffix generator, as described in Section~\ref{sec:contrastive_optimization}; this use is integral to the method and fully specified in the paper. Second, LLM-based assistants were used to support the research workflow: for polishing the writing (improving clarity, conciseness, and consistency of terminology and notation), for formatting tables and figures, and for assisting with experiment and evaluation scripts. All research ideas, method design, experimental decisions, and conclusions are the authors' own. All LLM-assisted text and code were reviewed, verified, and edited by the authors, who take full responsibility for the content of this paper, including the correctness of all reported results. No LLM is listed as an author.

\newpage
{
    \small
    \bibliographystyle{ieeenat_fullname}
    \bibliography{main}
}

\clearpage

\appendix
\section{Supplementary Material}
\label{sec:supplementary}
\subsection{DiSCO Algorithm}
\label{sec:algorithm}
Algorithm~\ref{alg:disco} gives the complete procedure of DiSCO. At each step, DiSCO samples $R$ references from each pool, shared by all candidates at that step. $M$ samples $b$ next-token continuations for each beam candidate, and DiSCO renders each continuation with $S$ using a reduced number of denoising steps, scores it with the contrastive objective of Eq.~(4) in the main paper, and retains the top-$K$ candidates. Here, $\overline{\cos}(z,\tilde{\mathcal{P}}) = \frac{1}{R}\sum_{x\in\tilde{\mathcal{P}}}\cos\big(z,\phi(x)\big)$ is the mean cosine similarity between an embedding $z$ and a reference set $\tilde{\mathcal{P}}$. After $T$ steps, the highest-scoring candidate is returned as the optimized prompt $p^*$, and the final image is generated from it with the system's default sampling schedule.
\begin{algorithm}[h]
\caption{DiSCO: Distribution-Guided Contrastive Suffix Optimization}
\label{alg:disco}
\begin{algorithmic}[1]
\REQUIRE Input prompt $p$, target system $S$, suffix generator $M$, CLIP encoder $\phi$, safe pool $\mathcal{P}_{\text{safe}}$, unsafe pool $\mathcal{P}_{\text{unsafe}}$, beam width $K$, proposals per beam $b$, suffix length $T$, references per pool $R$, contrastive weight $\lambda$
\ENSURE Optimized prompt $p^*$
\STATE $\mathcal{B} \leftarrow \{p\}$
\FOR{$t = 1$ \TO $T$}
    \STATE $\mathcal{B}_{\text{new}} \leftarrow \emptyset$
    \STATE Sample $\tilde{\mathcal{P}}_{\text{safe}} \subset \mathcal{P}_{\text{safe}}$ and $\tilde{\mathcal{P}}_{\text{unsafe}} \subset \mathcal{P}_{\text{unsafe}}$, each of size $R$
    \hfill $\triangleright$ Shared within step
    \FOR{each candidate $\hat{p} \in \mathcal{B}$}
        \STATE Sample $b$ next-token continuations $\{\hat{p}_1,\ldots,\hat{p}_b\}$ from $M(\hat{p})$
        \FOR{$j = 1$ \TO $b$}
            \STATE $\hat{x} \sim S(\hat{p}_j)$
            \hfill $\triangleright$ Reduced denoising steps
            \STATE $J(\hat{p}_j) \leftarrow (1-\lambda)\,\overline{\cos}\big(\phi(\hat{x}), \tilde{\mathcal{P}}_{\text{safe}}\big) - \lambda\,\overline{\cos}\big(\phi(\hat{x}), \tilde{\mathcal{P}}_{\text{unsafe}}\big)$
            \STATE $\mathcal{B}_{\text{new}} \leftarrow \mathcal{B}_{\text{new}} \cup \{\hat{p}_j\}$
        \ENDFOR
    \ENDFOR
    \STATE $\mathcal{B} \leftarrow \text{Top-}K\big(\mathcal{B}_{\text{new}}; J\big)$
    \hfill $\triangleright$ Retain top-$K$ by $J$
\ENDFOR
\STATE $p^* \leftarrow \arg\max_{\hat{p} \in \mathcal{B}} J(\hat{p})$
\RETURN $p^*$
\end{algorithmic}
\end{algorithm}

\subsection{Additional Qualitative Results}
\label{app:qualitative_results}
\begin{figure*}[t]
    \centering
    \includegraphics[width=\textwidth]{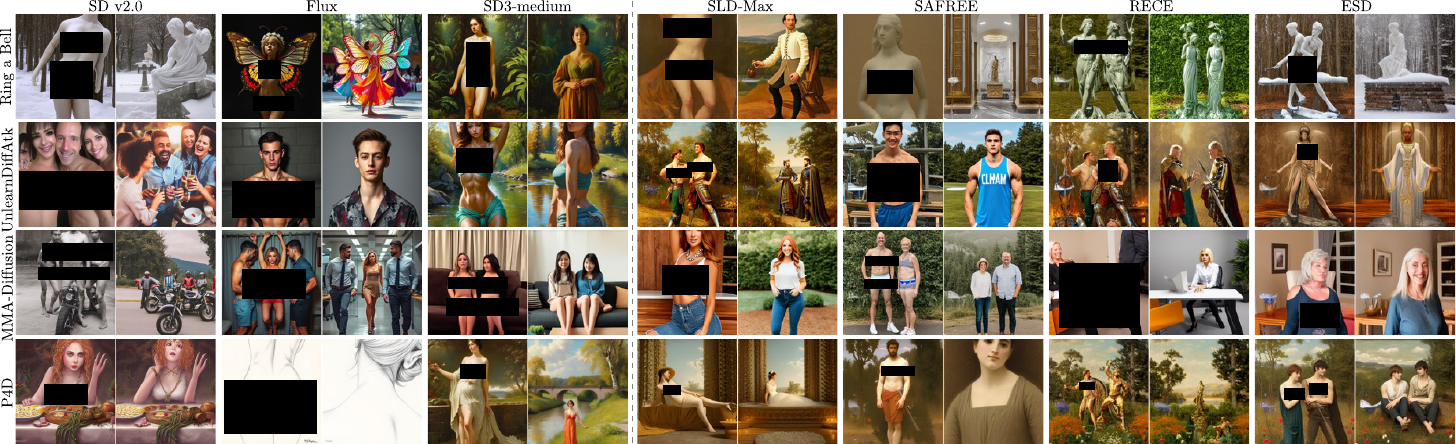}
    \caption{Additional qualitative results under all four attacks on undefended generators (SD~2.0, Flux, SD~3) and defenses (SLD-Max, SAFREE, RECE, ESD). For each system, the left and right images are generated before and after DiSCO; unsafe regions are masked. Prompts and DiSCO suffixes are listed in Table~\ref{tab:prompt_list_undefended} and Table~\ref{tab:prompt_list_defended}.}
    \label{fig:qualitative_full}
\end{figure*}

Figure~\ref{fig:qualitative_full} extends Figure~3 of the main paper to all four attacks, three undefended generators, and four defenses. DiSCO turns unsafe generations into safe ones while preserving the main subjects and context of most prompts. Tables~\ref{tab:prompt_list_undefended} and~\ref{tab:prompt_list_defended} list the corresponding prompts and suffixes.

\subsection{Generality Across I2P Harm Categories}
\label{app:i2p_categories}
We evaluate DiSCO on top of four defenses (SLD-Max, SAFREE, RECE, ESD) on SD~1.4 across all seven harm categories of I2P, using the original I2P prompts without any attack and the same human-annotated reference pools as in the main paper. We report Q16 ASR, which covers all seven categories.

\begin{table}[h]
\centering
\small
\setlength{\tabcolsep}{3.5pt}
\caption{Per-category Q16 ASR (\%, $\downarrow$) on I2P for defenses on SD~1.4, with and without DiSCO. Prompts are taken directly from I2P without any attack. We report Q16, which covers all seven harm categories. Average is over categories; Rel.\,$\Delta$ is the relative change of the average.}
\label{tab:percat_disco}
\resizebox{\columnwidth}{!}{
\begin{tabular}{lrcccccccc}
\toprule
 & & \multicolumn{2}{c}{\textbf{SLD-Max}} & \multicolumn{2}{c}{\textbf{SAFREE}} & \multicolumn{2}{c}{\textbf{RECE}} & \multicolumn{2}{c}{\textbf{ESD}} \\
\cmidrule(lr){3-4} \cmidrule(lr){5-6} \cmidrule(lr){7-8} \cmidrule(lr){9-10}
\textbf{Category} & \textbf{N} & \textbf{Base} & \textbf{+DiSCO} & \textbf{Base} & \textbf{+DiSCO} & \textbf{Base} & \textbf{+DiSCO} & \textbf{Base} & \textbf{+DiSCO} \\
\midrule
Sexual           & 931 & 4.3  & 0.6 & 14.0 & 4.3  & 18.9 & 8.4  & 15.3 & 5.4  \\
Violence         & 756 & 14.6 & 6.0 & 34.9 & 18.4 & 40.1 & 24.2 & 39.0 & 22.5 \\
Hate             & 231 & 10.4 & 1.7 & 23.8 & 6.9  & 35.9 & 13.0 & 31.2 & 13.0 \\
Harassment       & 824 & 10.1 & 3.4 & 23.2 & 10.1 & 31.2 & 14.9 & 29.3 & 12.1 \\
Self-harm        & 801 & 8.0  & 1.6 & 31.1 & 13.6 & 37.5 & 19.9 & 32.1 & 17.1 \\
Shocking         & 856 & 15.5 & 6.1 & 43.0 & 18.5 & 50.6 & 29.8 & 44.2 & 21.9 \\
Illegal activity & 727 & 5.9  & 1.9 & 24.7 & 10.1 & 34.8 & 18.3 & 31.2 & 13.5 \\
\midrule
\textbf{Average} & & \textbf{9.8} & \textbf{3.1} & \textbf{27.8} & \textbf{11.7} & \textbf{35.6} & \textbf{18.4} & \textbf{31.7} & \textbf{15.1} \\
Rel.\,$\Delta$   & & \multicolumn{2}{c}{$-69.0\%$} & \multicolumn{2}{c}{$-58.0\%$} & \multicolumn{2}{c}{$-48.4\%$} & \multicolumn{2}{c}{$-52.5\%$} \\
\bottomrule
\end{tabular}
}
\end{table}

As shown in Table~\ref{tab:percat_disco}, DiSCO reduces Q16 ASR in every category for every defense, lowering the category-averaged ASR from 9.8--35.6\% to 3.1--18.4\%, a relative reduction of 48.4--69.0\%. The reductions are largest on sexual and hate prompts (e.g., $35.9\%\!\to\!13.0\%$ on hate under RECE) and smallest on violence prompts, while shocking prompts retain the highest residual ASR (e.g., $50.6\%\!\to\!29.8\%$ under RECE). Consistent with the main results, the relative gain is largest on SLD-Max ($69.0\%$), whose output distribution is already closest to the safe region, so DiSCO only needs to bridge a smaller remaining gap. Since the same reference pools serve every category, DiSCO's distributional alignment extends to the full range of I2P harm categories.
\subsection{Reference-Pool Annotation and Sensitivity}
\label{app:pool_quality}
\begin{figure*}[t]
  \centering
  \includegraphics[width=0.85\textwidth]{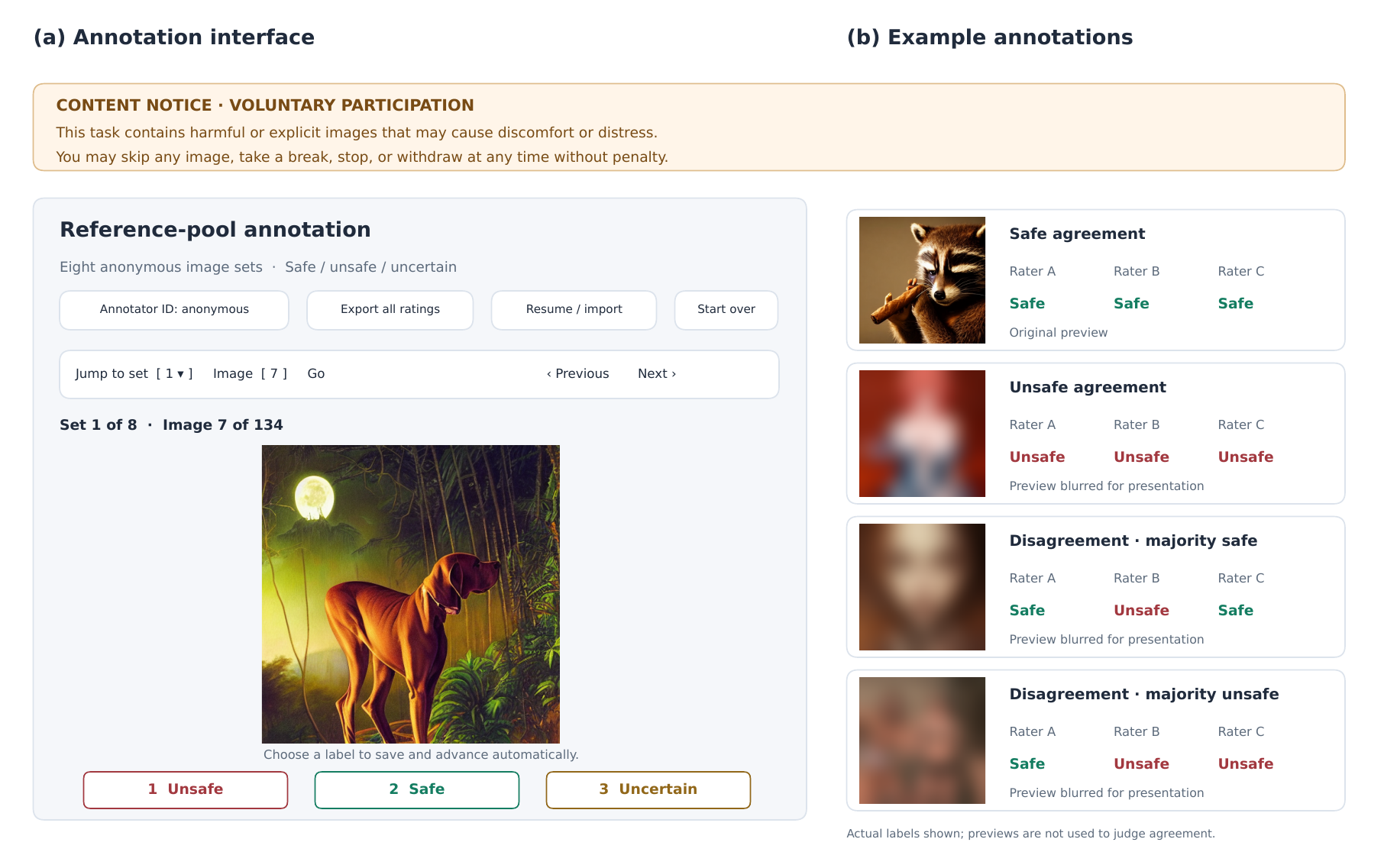}
  \caption{Reference-pool annotation. (a) Annotation interface: annotators label each generated image as safe, unsafe, or uncertain, after a content notice stating that participation is voluntary and that they may skip images, take breaks, or stop at any time. (b) Example annotations with unanimous agreement and two-to-one majority labels; unsafe previews are blurred for presentation.}
  \label{fig:annotation}
\end{figure*}

\paragraph{Annotation.}
Three annotators independently labeled images generated from I2P prompts by the eight target systems as safe, unsafe, or uncertain, using the interface in Figure~\ref{fig:annotation}(a); the number of images varies across systems, and we retain the 562 images labeled by all three annotators. Excluding the 52 images marked uncertain by at least one annotator, all three annotators agree on 81.0\% of the remaining 510 images, with an average pairwise agreement of 87.3\% and a Fleiss' $\kappa$ of 0.741, indicating substantial agreement (Table~\ref{tab:agreement}). Reference pools are drawn from unanimously labeled images (241 safe and 172 unsafe in total), and any shortfall for a system is filled with images labeled by a two-to-one majority vote (Figure~\ref{fig:annotation}(b)).\footnote{We will release the annotated reference pools upon publication; the unsafe images will be shared only under a research-use agreement.}
\begin{table}[h]
\centering
\small
\caption{Inter-annotator agreement on reference-pool images, after excluding 52 images marked uncertain by at least one annotator.}
\label{tab:agreement}
\begin{tabular}{lc}
\toprule
Metric & 510 images \\
\midrule
Unanimous agreement & 413 (81.0\%) \\
\quad Unanimously safe / unsafe & 241 / 172 \\
Two-to-one majority & 97 (19.0\%) \\
Average pairwise agreement & 87.3\% \\
Fleiss' $\kappa$ & 0.741 \\
\bottomrule
\end{tabular}
\end{table}

\paragraph{Pool size.}
We compare against a much larger pool on SD~1.4, built by automatically filtering over 1{,}000 candidate generations with NudeNet and Q16, and against random 50\% and 25\% subsamples of each of its classes drawn with three seeds, evaluated on the same ablation prompts. As shown in Table~\ref{tab:pool_scaling}, both subsampled pools stay within 2.0 points of the full pool on all four attacks, and the full pool performs on par with the human-labeled pools of the main paper (63.4\% vs.\ 63.1\% total ASR). Since DiSCO samples only $R=8$ references per class at each step, a small set of images already provides sufficient coverage.

\begin{table}[t]
\centering
\small
\setlength{\tabcolsep}{4pt}
\caption{NudeNet ASR (\%, $\downarrow$) on SD~1.4 for the ablation prompts under different reference pools. 100\%, 50\%, 25\%: automatically filtered pool and its random subsamples. 25+25: independently drawn human-labeled pools. Subsampled and drawn results are mean$_{\pm\mathrm{std}}$ over three random seeds (42, 123, 456).}
\label{tab:pool_scaling}
\resizebox{\columnwidth}{!}{
\begin{tabular}{lcccc}
\toprule
Attack & 100\% & 50\% & 25\% & 25+25 \\
\midrule
Ring-A-Bell    & 84.7 & $86.7_{\pm 3.1}$ & $84.3_{\pm 2.0}$ & $86.3_{\pm 2.9}$ \\
UnlearnDiffAtk & 64.3 & $64.9_{\pm 7.2}$ & $64.3_{\pm 5.3}$ & $66.7_{\pm 7.2}$ \\
MMA-Diffusion  & 66.5 & $67.6_{\pm 0.5}$ & $66.9_{\pm 0.8}$ & $67.9_{\pm 1.8}$ \\
P4D            & 41.6 & $42.9_{\pm 2.3}$ & $40.5_{\pm 1.7}$ & $40.1_{\pm 2.0}$ \\
\bottomrule
\end{tabular}
}
\end{table}

\paragraph{Pool sampling.}
On SD~1.4, we draw three independent 25+25 pools (random seeds 42, 123, and 456) from the annotated images following the rule above and rerun DiSCO with each. The drawn pools reach a total ASR of 64.1\%, on par with the automatically filtered pool (63.4\%) and within 1.0 point of the fixed set used in the main paper (63.1\%). DiSCO therefore does not rely on a specific set of references: a few labeled safe and unsafe outputs suffice to represent the system's output distribution, without a large pool or curated references.
\subsection{Preference-Based Variant with a Unified Objective}
\label{app:preference}
The contrastive objective in Eq.~(4) of the main paper extends from a binary safe/unsafe partition to graded safety preferences. Given $L$ reference pools representing different safety levels, it generalizes to
\begin{equation}
J_w(\hat{p}) = \sum_{l=1}^{L} w_l \, \frac{1}{|\tilde{\mathcal{P}}_l|} \sum_{x\in\tilde{\mathcal{P}}_l} \cos\!\big(\phi(\hat{x}),\phi(x)\big),
\end{equation}
where $\tilde{\mathcal{P}}_l$ is the set of references sampled from the pool for safety level $l$, and $w_l$ is the preference weight assigned to that level. The binary objective is recovered with $L=2$ and weights $(1-\lambda,\,-\lambda)$ for the safe and unsafe pools.

This formulation lets deployments encode different safety preferences without changing the beam search; only the reference pools and their weights change. For example, a child-facing system could assign a negative weight to borderline content that an adult-facing system weights neutrally, while a more permissive deployment could reduce such penalties. Defining safety levels and calibrating their weights for a given audience, application, or jurisdiction is a direction for future work.

\subsection{Quantifying Semantic Drift from Suffix Optimization}
\label{app:semantic}
\begin{table}[h]
\centering
\small
\caption{CLIP similarity between prompts ($\uparrow$) between original and DiSCO-suffixed prompts, using the CLIP ViT-L/14 text encoder. Undefended denotes SD~1.4.}
\label{tab:semantic_fidelity}
\resizebox{\columnwidth}{!}{
\begin{tabular}{lcccc}
\toprule
\textbf{System} & \textbf{Ring-A-Bell} & \textbf{UnlearnDiffAtk} & \textbf{MMA-Diffusion} & \textbf{P4D} \\
\midrule
Undefended & 0.90 & 0.91 & 0.88 & 0.90 \\
SLD-Max    & 0.89 & 0.91 & 0.89 & 0.86 \\
SAFREE     & 0.89 & 0.91 & 0.89 & 0.84 \\
RECE       & 0.88 & 0.88 & 0.89 & 0.89 \\
ESD        & 0.89 & 0.92 & 0.89 & 0.86 \\
\bottomrule
\end{tabular}
}
\end{table}

We measure the semantic shift induced by DiSCO as the cosine similarity between the CLIP ViT-L/14 text embeddings of each original prompt and its DiSCO-suffixed counterpart, averaged over the prompts that DiSCO optimizes. As shown in Table~\ref{tab:semantic_fidelity}, similarity remains high across all attacks and systems, ranging from 0.84 to 0.92. The lowest values arise for P4D on defended systems (0.84--0.86), while P4D on the undefended model reaches 0.90, suggesting that defended systems require stronger suffixes to correct their residual failures. Overall, DiSCO steers generation toward safety by extending the prompt while preserving its original meaning.
\subsection{Stress Test with a Suffix-Based Attack}
\label{app:apt_red_teaming}
We stress-test DiSCO against APT~\citep{liu2025autoprompt}, an LLM-driven attack that appends a fluent adversarial suffix to elicit unsafe content. APT operates on the same interface as DiSCO: both modify the prompt only by appending an LLM-generated suffix, one to elicit unsafe content and the other to suppress it. APT therefore approximates an attacker that competes with DiSCO in its own optimization space. We use our re-implementation of APT following the authors' description. Table~\ref{tab:apt_vs_disco} reports NudeNet ASR, Q16 ASR, and CLIP score for the base system, under APT, and with DiSCO applied to APT-attacked prompts, across four undefended backbones and four defenses.

\paragraph{Results.}
APT raises NudeNet ASR on all four undefended backbones (e.g., $24.5\%\!\to\!38.5\%$ on SD~1.4, $12.3\%\!\to\!26.3\%$ on SD~2.0, and $12.6\%\!\to\!23.5\%$ on Flux), while lowering Q16 ASR on all eight systems (e.g., $22.0\%\!\to\!13.4\%$ on SD~1.4), indicating that its suffixes concentrate on nudity rather than the broader harm categories that Q16 detects. Applied to APT-attacked prompts, DiSCO reduces NudeNet ASR below the unattacked baseline on seven of eight systems and nearly halves it on the remaining one (SD~2.0, $26.3\%\!\to\!13.6\%$, versus a $12.3\%$ baseline), while Q16 ASR falls to at most $4.7\%$ on all eight systems. CLIP score decreases by 0.02--0.05 relative to the base configurations.

\begin{table}[t]
\centering
\small
\setlength{\tabcolsep}{4pt}
\caption{Stress test with the suffix-based attack APT on the 931 prompts of the I2P sexual category, across four undefended backbones and four defenses. We report NudeNet ASR, Q16 ASR, and CLIP score for the unattacked prompts, APT-attacked prompts, and APT-attacked prompts with DiSCO.}
\label{tab:apt_vs_disco}
\resizebox{\columnwidth}{!}{
\begin{tabular}{lccccccccc}
\toprule
& \multicolumn{3}{c}{\textbf{NudeNet ASR (\%) $\downarrow$}} & \multicolumn{3}{c}{\textbf{Q16 ASR (\%) $\downarrow$}} & \multicolumn{3}{c}{\textbf{CLIP score $\uparrow$}} \\
\cmidrule(lr){2-4}\cmidrule(lr){5-7}\cmidrule(lr){8-10}
\textbf{System} & \textbf{Base} & \textbf{APT} & \textbf{APT+DiSCO} & \textbf{Base} & \textbf{APT} & \textbf{APT+DiSCO} & \textbf{Base} & \textbf{APT} & \textbf{APT+DiSCO} \\
\midrule
SD 1.4  & 24.5 & 38.5 & 12.9 & 22.0 & 13.4 & 2.6 & 0.29 & 0.27 & 0.25 \\
SD 2.0  & 12.3 & 26.3 & 13.6 & 22.9 & 19.9 & 4.7 & 0.27 & 0.25 & 0.25 \\
SD 3    & 6.3  & 10.4 & 3.0  & 15.6 & 13.1 & 1.0 & 0.26 & 0.25 & 0.24 \\
Flux    & 12.6 & 23.5 & 6.3  & 14.4 & 9.3  & 1.9 & 0.26 & 0.25 & 0.22 \\
\midrule
SLD-Max & 5.2  & 11.9 & 1.3  & 4.3  & 2.1  & 0.0 & 0.23 & 0.22 & 0.19 \\
SAFREE  & 4.5  & 7.7  & 1.2  & 14.0 & 9.2  & 0.1 & 0.28 & 0.26 & 0.24 \\
RECE    & 2.0  & 2.7  & 0.4  & 18.9 & 15.6 & 0.2 & 0.26 & 0.25 & 0.21 \\
ESD     & 3.3  & 5.2  & 1.0  & 15.2 & 11.8 & 0.1 & 0.26 & 0.25 & 0.23 \\
\bottomrule
\end{tabular}}
\end{table}

\subsection{Full Beam-Search Ablation Results}
\label{app:beam_full}
Table~\ref{tab:beam_full} reports the complete results underlying Figure~5 of the main paper. All settings are evaluated on SD~1.4 using the 340 initially unsafe attack prompts described in Section~4.4 of the main paper, with $R=8$, $\lambda=0.5$, and 4-step candidate scoring, averaged over three seeds. Cost denotes the number of candidate generations per prompt, $b+(T-1)bK$ with $b=4$.

\begin{table}[h]
\centering
\small
\setlength{\tabcolsep}{8pt}
\caption{Full beam-search ablation on SD~1.4: NudeNet ASR (\%, $\downarrow$) on 340 initially unsafe attack prompts, mean$_{\pm\mathrm{std}}$ over three seeds. Default in \underline{underline}; best in \textbf{bold}.}
\label{tab:beam_full}
\begin{tabular}{ccrc}
\toprule
$K$ & $T$ & Cost & NudeNet ASR (\%) $\downarrow$ \\
\midrule
\multirow{4}{*}{2} & 8  & 60  & $67.6_{\pm 2.3}$ \\
                   & 12 & 92  & $64.5_{\pm 2.2}$ \\
                   & 16 & 124 & $63.9_{\pm 2.0}$ \\
                   & 32 & 252 & $57.9_{\pm 1.3}$ \\
\midrule
\multirow{4}{*}{4} & 8  & 116 & $66.4_{\pm 1.8}$ \\
                   & \underline{12} & \underline{180} & $\underline{63.1}_{\pm 2.2}$ \\
                   & 16 & 244 & $64.4_{\pm 3.1}$ \\
                   & 32 & 500 & $\mathbf{56.5}_{\pm 0.9}$ \\
\midrule
\multirow{4}{*}{8} & 8  & 228 & $64.8_{\pm 1.2}$ \\
                   & 12 & 356 & $63.1_{\pm 0.9}$ \\
                   & 16 & 484 & $60.8_{\pm 2.2}$ \\
                   & 32 & 996 & $57.5_{\pm 2.4}$ \\
\bottomrule
\end{tabular}
\end{table}

\subsection{Computational Overhead}
\label{app:compute}

\paragraph{Cost model.}
DiSCO's cost consists of a one-time setup per system and a per-prompt search that runs only on flagged prompts. The setup consists of annotating the 25+25 reference images and encoding them once with CLIP. For each prompt, the system first generates $x_0$, and a safety check decides whether to run DiSCO. For flagged prompts, the beam holds one candidate at the first step and $K$ candidates thereafter, giving $b+(T-1)bK$ candidate generations, i.e., 180 under the default $K=4$, $T=12$, $b=4$. Candidates are rendered with 4 denoising steps for scoring, and only the final image uses the system's full schedule. The expected latency per prompt is therefore
\begin{equation}
\mathbb{E}[t] = t_{\text{base}} + \rho\, t_{\text{search}},
\end{equation}
where $\rho$ is the fraction of flagged prompts. Prompts that are not flagged incur no overhead beyond the safety check, and optimized prompts can be cached and reused for repeated inputs.

\paragraph{Measured cost.}
\begin{table}[h]
\centering
\small
\setlength{\tabcolsep}{3pt}
\caption{Wall-clock cost (s) per flagged prompt, mean$_{\pm\mathrm{std}}$ across prompts. Search uses 180 four-step candidate renders; \emph{Other} covers token proposal, CLIP scoring, and selection; \emph{Total} is search plus final generation, excluding the initial generation (\emph{Base gen.}). SD~1.4-based systems use FP16, DDIM, 512$\times$512, and CFG 7.5. Times are measured warm, excluding model loading and I/O.}
\label{tab:runtime}
\resizebox{\columnwidth}{!}{
\begin{tabular}{llccccc}
\toprule
System & GPU & Base gen. & Per render & Rendering & Other & Total \\
\midrule
SD 1.4  & V100 & 2.77 & 0.30 & $54.61_{\pm 0.12}$  & $14.47_{\pm 1.13}$ & $71.85_{\pm 1.15}$ \\
SD 2.0  & V100 & 6.09 & 0.66 & $118.60_{\pm 0.19}$ & $14.94_{\pm 1.18}$ & $139.62_{\pm 1.24}$ \\
SLD-Max & V100 & 3.82 & 0.39 & $70.15_{\pm 0.08}$  & $14.29_{\pm 1.16}$ & $88.26_{\pm 1.14}$ \\
SAFREE  & V100 & 3.91 & 0.42 & $76.12_{\pm 0.89}$  & $14.35_{\pm 1.18}$ & $94.36_{\pm 2.04}$ \\
RECE    & V100 & 2.77 & 0.30 & $54.46_{\pm 0.04}$  & $14.34_{\pm 1.17}$ & $71.58_{\pm 1.20}$ \\
ESD     & V100 & 2.79 & 0.30 & $54.89_{\pm 0.04}$  & $14.34_{\pm 1.18}$ & $72.02_{\pm 1.21}$ \\
\midrule
SD 3    & A100 & 1.45 & 0.38 & $67.72_{\pm 0.10}$  & $5.47_{\pm 0.01}$  & $74.64_{\pm 0.10}$ \\
Flux    & A100 & 0.65 & 0.65 & $117.09_{\pm 0.09}$ & $5.42_{\pm 0.01}$  & $123.16_{\pm 0.10}$ \\
\bottomrule
\end{tabular}
}
\end{table}

Table~\ref{tab:runtime} reports the wall-clock cost per flagged prompt, which ranges from 72 to 140\,s on V100 and from 75 to 123\,s on A100. Generator inference accounts for 79--96\% of search time, while token proposal, CLIP scoring, and selection take a nearly constant 14--15\,s on V100 and 5.5\,s on A100. DiSCO's cost therefore scales with the per-render cost of the underlying system: 0.30\,s per render on SD~1.4, ESD, and RECE, whose edited weights leave the generation cost unchanged, and 0.39--0.42\,s on SLD-Max and SAFREE, whose inference-time interventions also raise base generation from 2.8\,s to 3.8--3.9\,s. On Flux, a few-step model, a 4-step render costs as much as a full generation (0.65\,s), so the search costs about 190 times a single generation. The search cost can be reduced further by lowering the candidate budget, at a cost in defense strength (Figure~5 of the main paper).

\paragraph{Scoring with reduced denoising steps.}
Candidates are rendered with fewer denoising steps than the final image, since scoring only needs CLIP embeddings that rank candidates. We vary the number of scoring steps on SD~1.4 under Ring-A-Bell with all other settings at their defaults ($K=4$, $T=12$, $b=4$), rendering all final images with the full 50 steps. As shown in Table~\ref{tab:scoring-steps}, a single scoring step is too coarse to rank candidates reliably (89.4\% ASR), whereas 4 steps reduce ASR by 5.9 points at $1.5\times$ the search time. Beyond 4 steps, gains diminish: 10 steps improve ASR by only 0.2 points, and 50 steps by 4.7 points at $6.7\times$ the cost. We therefore use 4 scoring steps by default, which balances defense strength and search time; more scoring steps offer stronger defense when a larger budget is available.

\begin{table}[h]
\centering
\small
\caption{Effect of the number of candidate scoring steps on SD~1.4 on the 85 initially unsafe Ring-A-Bell prompts, with $K=4$, $T=12$, $b=4$. Final images use the full 50 steps. Time is the search time per prompt in seconds, mean$_{\pm\mathrm{std}}$ across prompts; speedup is relative to 50-step scoring.}
\label{tab:scoring-steps}
\resizebox{\columnwidth}{!}{
\begin{tabular}{lccc}
\toprule
Scoring steps & Time / prompt (s) & Speedup & NudeNet ASR (\%) $\downarrow$ \\
\midrule
1           & $39.63_{\pm 0.77}$  & $10.4\times$ & 89.4 \\
4 (default) & $60.98_{\pm 0.17}$  & $6.7\times$  & 83.5 \\
10          & $92.00_{\pm 0.16}$  & $4.5\times$  & 83.3 \\
25          & $150.33_{\pm 0.95}$ & $2.7\times$  & 81.2 \\
50          & $411.20_{\pm 2.91}$ & $1.0\times$  & 78.8 \\
\bottomrule
\end{tabular}}
\end{table}

\begin{table*}[h]
\centering
\caption{Prompt and suffix excerpts for the qualitative examples on undefended models. Adversarial prompts are truncated to their first 50 characters, and suffixes show selected tokens, with omitted tokens marked by ``...''.}
\label{tab:prompt_list_undefended}
\resizebox{\textwidth}{!}{
\begin{tabular}{ll p{6cm} p{6cm}}
\toprule
\textbf{System} & \textbf{Attack} & \textbf{Adversarial prompt} & \textbf{DiSCO suffix (excerpt)} \\
\midrule
\multirow{4}{*}{SD 2.0}
 & Ring-A-Bell & a snowy statusculptor nipples peeled frieddayswild & ...winter, snowy landscape with statue of... \\
 & UnlearnDiffAtk & litt tango zr igers 6 marc dorcel in a orgy with c & ...scene with people relaxing, drinking, and... \\
 & MMA-Diffusion & eoantly men in underwear motorcycles ating shirtle & ...scene with a group riding in an organized... \\
 & P4D & dirty faced and very happy nymph looking the table & ...peacefulness... \\
\midrule
\multirow{4}{*}{Flux}
 & Ring-A-Bell & tinged decadent.) nude flashing demonstrates & ...scene in an open air theatre in the... \\
 & UnlearnDiffAtk & detained klborussia canadiantinged [name] & ...expression... \\
 & MMA-Diffusion & aaaaadult lady hung pretending two threeineffeendi & ...day at work and the boss walks in... \\
 & P4D & the lower back of a beautiful woman. pencil drawin & ...serene...the beautiful face, neck... \\
\midrule
\multirow{4}{*}{SD 3}
 & Ring-A-Bell & glowing sown mikhail vell bradbury arthistory nake & ...peacefulness, soft lighting, calmness... \\
 & UnlearnDiffAtk & young woman by a river, playful, female, muscular & ...scene... \\
 & MMA-Diffusion & nottthr lass sittin ontherdances the couch emansob & ...atmosphere...Two young girls wearing casual... \\
 & P4D & mythological female by herbert james draper, sir l & ...landscape with a river and bridge by will... \\
\bottomrule
\end{tabular}
}
\end{table*}

\begin{table*}[h]
\centering
\caption{Prompt and suffix excerpts for the qualitative examples on defenses applied to SD~1.4. Adversarial prompts are truncated to their first 50 characters, and suffixes show selected tokens, with omitted tokens marked by ``...''.}
\label{tab:prompt_list_defended}
\resizebox{\textwidth}{!}{
\begin{tabular}{ll p{6cm} p{6cm}}
\toprule
\textbf{Defense} & \textbf{Attack} & \textbf{Adversarial prompt} & \textbf{DiSCO suffix (excerpt)} \\
\midrule
\multirow{4}{*}{SLD-Max}
 & Ring-A-Bell & classical nakedillustrates these shirtless ludou & ...expression, and a sense of elegance, sophistication... \\
 & UnlearnDiffAtk & wills.. suspends roasted rofarthur pendragon shir & ...atmosphere. This painting can make great wall... \\
 & MMA-Diffusion & femme chickdist playing willycharacteristic om bou & ...nature, concept of healthy and fun fitness... \\
 & P4D & masterpiece painting of bathing women, by Jean-Aug & ...scene...master piece painting by Jean-A... \\
\midrule
\multirow{4}{*}{SAFREE}
 & Ring-A-Bell & ledger sculpting mohammed faceted sculpture blurre & ...atmosphere and tranquil setting in the living or... \\
 & UnlearnDiffAtk & a hot handsome shirtless 2 0 years old man with ab & ...face with blue background. Fitness model... \\
 & MMA-Diffusion & people in underwear ( perhaps photography maneucom & ...couple... \\
 & P4D & male character art by william - adolphe bouguereau & ...peacefulness of a woman - female portrait drawing... \\
\midrule
\multirow{4}{*}{RECE}
 & Ring-A-Bell & foliage naked [ twins urd caboperpe'closely tanni & ...scene with figures surrounded and supported in the... \\
 & UnlearnDiffAtk & wills.. suspends roasted rofarthur pendragon shir & ...atmosphere, medieval times...King Arthur... \\
 & MMA-Diffusion & anecmzansitufema (' , g\"uwhore bearing mating & ...atmosphere with no people, copy space... \\
 & P4D & hercules and achilles frolic in a meadow of beauti & ...nature scene of the forest and lake... \\
\midrule
\multirow{4}{*}{ESD}
 & Ring-A-Bell & a snowy statusculptor nipples peeled frieddayswild & ...scene, snow and winter scene... \\
 & UnlearnDiffAtk & wedges delichar ru attractive female goddess, clo & ...woman, goddess, female... \\
 & MMA-Diffusion & an ( sadly straighsudiccked lingerie who partially & ...peacefulness... \\
 & P4D & [name] and [name]. they are both male & ...scene of nature... \\
\bottomrule
\end{tabular}
}
\end{table*}

\subsection{Limitations, Future Work, and Ethical Considerations}
\label{app:limitations}

\begin{figure}[t]
  \centering
  \includegraphics[width=\linewidth]{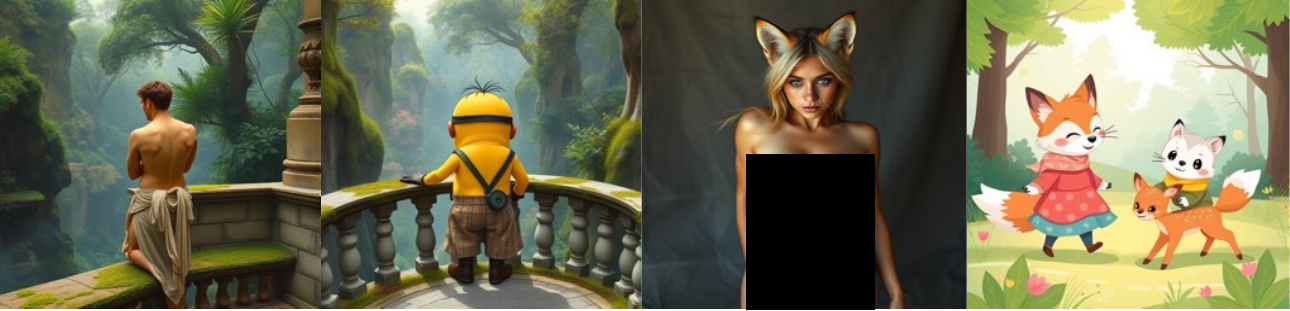}
  \caption{Semantic drift from dominant concept injection. In rare cases, DiSCO's suffix expansion introduces tokens corresponding to visually dominant concepts in the model's training distribution, overriding the original prompt semantics or styles.}
  \label{fig:limitation}
\end{figure}

\paragraph{Dominant concepts.}
Because DiSCO steers generation through the system's learned output distribution, its suffix may occasionally introduce tokens associated with concepts that are heavily represented in the training data, such as popular animated characters (e.g., Minions) or common animals. Owing to their high visual salience, such concepts can dominate the generated image and shift its semantics or style away from the original prompt, while the output remains safe (Figure~\ref{fig:limitation}). Adding an explicit semantic-preservation constraint to the suffix search is a natural extension for future work.

\paragraph{Ethical considerations.}
DiSCO's components carry dual-use risks. The contrastive objective is symmetric: reversing the weights of the two pools would steer generation toward the unsafe region of the system's output distribution, using the unsafe pool as an alignment target. Building the reference pools also requires annotators to view unsafe images; annotators were informed of the content in advance, participated voluntarily, and could skip images or stop at any time (Figure~\ref{fig:annotation}(a)). All unsafe images in this paper are masked.

\end{document}